\documentclass[preprint,12pt]{elsarticle}

\usepackage{amsmath,amssymb,amsfonts}
\usepackage{algorithmic}
\usepackage{graphicx}
\usepackage{textcomp}
\usepackage{float}
\usepackage{multirow}

\journal{Computers in Biology and Medicine}
\begin{document}
\begin{frontmatter}
\title{Prompt-enhanced Hierarchical Transformer Elevating Cardiopulmonary Resuscitation Instruction via Temporal Action Segmentation}

\author[1]{Yang Liu}
\author[1,2]{Xiaoyun Zhong}
\author[1,2]{Shiyao Zhai}
\author[1,2]{Zhicheng Du}
\author[3]{Zhenyuan Gao}
\author[4]{Qiming Huang}
\author[1]{Canyang Zhang}
\author[5]{Bin Jiang}
\author[1,2]{Vijay Kumar Pandey}
\author[1]{Sanyang Han}
\author[1]{Runming Wang}
\author[6]{Yuxing Han}
\author[1,2]{Peiwu Qin\corref{cor1}}
\ead{pwqin@sz.tsinghua.edu.cn}

\cortext[cor1]{Corresponding author.}

\address[1]{Insititute of Biopharmaceutics and Health Engineering, Tsinghua Shenzhen International Graduate School, Shenzhen, 518055, China}
\address[2]{Center of Precision Medicine and Healthcare, Tsinghua-Berkeley Shenzhen Institute, Shenzhen, 518055, China}
\address[3]{Miragestars Inc., Tianjin, 300392, China}
\address[4]{Shenzhen ZNV Technology Co., Ltd, Shenzhen, 518057, China}
\address[5]{School of Mechanical, Electrical \& Information Engineering, Shandong University, Weihai 264209, China}
\address[6]{Tsinghua Shenzhen International Graduate School, Shenzhen, 518055, China}

\begin{abstract}
The vast majority of people who suffer unexpected cardiac arrest are performed cardiopulmonary resuscitation (CPR) by passersby in a desperate attempt to restore life, but endeavors turn out to be fruitless on account of disqualification. Fortunately, many pieces of research manifest that disciplined training will help to elevate the success rate of resuscitation, which constantly desires a seamless combination of novel techniques to yield further advancement. To this end, we collect a custom CPR video dataset in which trainees make efforts to behave resuscitation on mannequins independently in adherence to approved guidelines, thereby devising an auxiliary toolbox to assist supervision and rectification of intermediate potential issues via modern deep learning methodologies. Our research empirically views this problem as a temporal action segmentation (TAS) task in computer vision, which aims to segment an untrimmed video at a frame-wise level. Here, we propose a Prompt-enhanced hierarchical Transformer (PhiTrans) that integrates three indispensable modules, including a textual prompt-based Video Features Extractor (VFE), a transformer-based Action Segmentation Executor (ASE), and a regression-based Prediction Refinement Calibrator (PRC). The backbone of the model preferentially derives from applications in three approved public datasets (GTEA, 50Salads, and Breakfast) collected for TAS tasks, which accounts for the excavation of the segmentation pipeline on the CPR dataset. In general, we unprecedentedly probe into a feasible pipeline that genuinely elevates the CPR instruction qualification via action segmentation in conjunction with cutting-edge deep learning techniques. Associated experiments advocate our implementation with multiple metrics surpassing 91.0\%.
\end{abstract}

\end{frontmatter}

\section{Introduction}
\label{sec:introduction}
Out-of-hospital cardiac arrest (OHCA) is a universal public health issue undergone by about 3.8 million people annually, with only 8\% to 12\% surviving hospital discharge \cite{Brooks2022Optimizing}. Characterized as blood flow or breathing stops, OHCA induces permanent brain damage or death happens acutely. Performing cardiopulmonary resuscitation (CPR) could serve as an emergency procedure for OHCA, which maintains the blood flow and breathing until advanced medical help arrives \cite{sandroni2021brain}. There has been a large volume of research and practice for decades to investigate CPR \cite{Association1974Standards,Lancet2018Out,HasselqvistAx2015Early}. In 1891, the first chest compression on a human being were performed by Friedrich Maass \cite{Lancet2018Out}. The first guidelines for CPR were released about 50 years ago \cite{Association1974Standards}. Despite a long history of deploying CPR against OHCA, survival remains dismally low. There are indications that CPR performance influences the outcome \cite{Wik1994Quality,Gallagher1995Effectiveness}. From this perspective, a great deal of research \cite{Pivac2020impact,Khanji2022Cardiopulmonary,Bielski2022Outcomes} seeks and proves the positive effects of CPR instruction. Besides, the increasing number of people suffering OHCA worldwide makes intensive CPR education even more imperative. To enable CPR education as a mandatory part of society, not only should we cultivate the awareness of social responsibility, but employ more comprehensive approaches \cite{Pivac2020impact}. Specifically, traditional assessment of CPR skills involves strenuous manual efforts, which lacks efficiency and repeatability \cite{Pivac2020impact,Khanji2022Cardiopulmonary}. Extensive attempts to revive those who sustain OHCA will be probably further improved with the combination of promising novel computer techniques and widespread application.

One most relevant specific research \cite{Alayrac2016Unsupervised} emerges by collecting a dataset of real-world instruction videos from the Internet, containing performing CPR and four non-medical tasks. However, the work intends to address the problem of automatically learning the main steps to complete a certain task, neither implemented particularly for formal CPR behaviors nor designed for CPR instruction. Given the aforementioned dataset, several subsequent works focus on action segmentation in an unsupervised \cite{Piergiovanni2021Unsupervised} or weakly supervised way \cite{Ding2022Temporal} without proposing concrete challenges pertaining to CPR instruction. The positive impacts of CPR action segmentation are not clearly defined yet. Therefore, we evoke one ensuing challenge: \emph{How better could we elevate CPR instruction with action segmentation?}

To this end, we resort to investigating a specific realm called temporal action segmentation (TAS), which has gradually developed into one of the high-profile research spotlights in computer vision. The universal goal of the TAS task is to identify activities in untrimmed videos at a frame-wise level. It has promoted a wealth of applications in human behavior analysis from video summarization \cite{Apostolidis2021Video}, video surveillance \cite{Vishwakarma2013survey}, action recognition \cite{Jhuang2013Towards,Kong2022Human}, to skill assessment \cite{Liu2021Towards}. With prosperous computing power reinforcement, understanding single-semantics short video clips has been gradually outmoded in the TAS task in favor of larger, more complex untrimmed videos \cite{Wang2020Boundary,Farha2019Ms}, which requires both intrinsic and extrinsic correlations of actions. Conventional segmentation methods \cite{Cheng2014Temporal} like Temporal Convolutional Networks (TCNs) consider single frames or short video segments for feature representation. They overlook the latent relationship among contextual actions, leading to poor performance, especially in long videos. Accordingly, some studies \cite{Huang2016Connectionist,Singh2016multi} exploit Recurrent Neural Networks (RNNs) to model each action clip to maintain local dependencies but still struggle to handle longer videos effectively due to the inherent spatio-temporal complexity of RNN. To equip the model with relational reasoning, methods utilizing Graph Convolutional Networks (GCNs) regard each action as a single node on the graph and edges represent the contextual relationship \cite{Wang2021Temporal,Huang2020Improving}. However, these preceding works all adopt frame-wise features extracted by pre-trained I3D \cite{Carreira2017Quo} network, which might not be adequate enough to excavate effective representations of videos. In the past few years, it can be witnessed that both transformer-based \cite{Vaswani2017Attention,Dosovitskiy2020image} and prompt-based architectures \cite{Brown2020Language,Radford2021Learning} have flourished in artificial intelligence, which tremendously lightens our research on addressing deficiencies of preceding methods. Particularly in visual applications, Transformer-based architectures hold the potential to integrate the information between sequential elements that are far from each other with powerful scalability. Prompt-based architectures could enhance the visual features with representative linguistic semantics.

Now we attempt to yield a feasible resolution for the proposed challenge: We preliminarily establish a custom CPR dataset involving videos of participants performing the whole process of CPR in a standard green screen laboratory environment. After that, we devise a crafted architecture to perform action segmentation. Specifically, we propose a Prompt-enhanced hierarchical Transformer (PhiTrans) that integrates three integral modules: 1) a textual prompt-based Video Features Extractor (VFE) module that extracts abundant frame-wise features; 2) a transformer-based Action Segmentation Executor (ASE) module that deduces the contextual relationship while adaptive to long frame sequences; 3) a regression-based Prediction Refinement Calibrator (PRC) module that further alleviates the over-segmentation issues highlighted in the TAS task. Finally, we observe the model performance on the custom CPR dataset and claim that it can serve as an auxiliary toolbox applied in action segmentation for assisting CPR instruction by automatically identifying potential omission, repetition, or out-of-order situations at a frame-wise level, allowing trainees to rectify the workflow in real-time free of experts.

Our main contributions are chronologically three-fold:
\begin{itemize}
    \item We collect a custom CPR dataset covering videos of the entire cardiopulmonary resuscitation evaluation performed by each participant, along with corresponding frame-level semantic annotation.
    \item We propose an integrated model, called PhiTrans, especially applied for CPR action segmentation, including three integral modules: Video Features Extractor, Action Segmentation Executor, and Prediction Refinement Calibrator.
    \item To the best of our knowledge, we unprecedentedly probe into a feasible pipeline that genuinely elevates the CPR instruction qualification via action segmentation in conjunction with cutting-edge deep learning techniques.
\end{itemize}

\section{Related Work}
\subsection{Temporal Action Segmentation}
Temporal action segmentation (TAS), one of the most challenging topics in advanced video comprehension, aims to extract frame-wise features from untrimmed videos and categorize them chronologically with pre-defined action labels. Action localization, video summarization, and other downstream applications benefit from products of the TAS task as input. Over the past decade, a multitude of methods \cite{Karaman2014Fast,Kuehne2016end,Bhattacharya2014Recognition,Cheng2014Temporal,Lea2017Temporal,zhang2018weighted} have leveraged models for action prediction with the extracted frame-wise features. Traditional paradigms involve sliding windows with non-maximum suppression \cite{Karaman2014Fast}, Hidden Markov Models (HMM) \cite{Kuehne2016end}, Linear Dynamical Systems (LDS) \cite{Bhattacharya2014Recognition}, and Bayesian Non-parametric Models (BNM) \cite{Cheng2014Temporal}.

These methods encounter common obstacles to modeling long-range dependencies. To alleviate this issue, RNNs \cite{Singh2016multi} and TCNs \cite{Lea2017Temporal} are deployed to capture global dependencies and the contextual information of adjacent frames. After that, Multi-Stage Temporal Convolutional Network (MS-TCN) \cite{Farha2019Ms} combines TCNs with multi-stage patterns to make remarkable progress in the TAS task, wherein plural stages are stacked to refine the predictions from the preceding output successively. In addition, other methods aim to model the TAS task from a unique perspective. Self-Supervised Temporal Domain Adaptation (SSTDA) \cite{Chen2020Action} trains with two auxiliary tasks of binary and sequential domain prediction. Dilated Temporal Graph Reasoning Module (DTGRM) \cite{Wang2021Temporal} builds multi-level dilated temporal graphs to simulate temporal dependencies between video frames at different timescales. These exceptive methods lose portability on account of complex task patterns. Considering the spatio-temporal complexity of CPR action segmentation and accessibility for downstream design, we follow the same philosophy of multi-stages for iterative refinement.

\subsection{Transformer-based Architecture}
Initially designed for Natural Language Processing (NLP) related tasks, Transformer \cite{Vaswani2017Attention} has motivated a tremendous leap forward in capabilities for pre-training on larger datasets and fine-tuning on smaller task-specific datasets with computational efficiency and scalability. The visual application \cite{Dosovitskiy2020image} of Transformer constantly challenges the dominant status of Convolutional Neural Networks (CNNs)\cite{zhang2022rcmnet,hassan2022neuro,hassan2023retinal,bhardwaj2022machine,guan2023prevalence,zhang2022ai,liu2022mixed,chen2021accelerated,chen2023crispr}. Vision Transformers require less vision-specific inductive bias and maintain more global information relying on the Multi-head Self-Attention (MSA).

Witnessed the success of Transformer in image classification, image segmentation, and other vision tasks, one study \cite{Yi2021Asformer} explores the transfer implementation of Transformer on the TAS task. The work presents major concerns underwent and accordingly proposes local connectivity inductive bias and hierarchical representation pattern, allowing vanilla Transformer scale to the TAS task. Such adaptions produce a hybrid Transformer architecture with temporal convolution included, which leads to fine-grained loss between adjacent frames as the depth layers increase. Most recently, follow-up works tend to improve this issue in two ways: establish a pure Transformer model \cite{Du2022Do} or refine temporal convolutions \cite{Wang2022Cross}. Although these efforts seem favorable, the robustness remains to be verified due to the lack of open sources. Therefore, our approach preserves the backbone and excavates promising performance with extensive optimization. Particularly, considering the resemblance among CPR actions, we enjoy a specific action segmentation refinement framework \cite{Ishikawa2021Alleviating}, laying the foundation for our endeavors to alleviate over-segmentation errors highlighted in the TAS task.

\subsection{Prompt-based Learning}
Served as an evolutionary group of Machine Learning (ML) model training approaches, prompt-based learning preliminarily allows people linguistically specify a certain task for the pre-trained Large Language Model (LLM) to compile and complete \cite{Brown2020Language}. To present a more intuitive perception of prompt-based learning, this section primarily introduces the identification of prompt and prompt engineering. Essentially, a prompt is an instruction depicted in natural language for the model to execute. The procedure of building the ideal prompt for a specific task is called prompt engineering. Subsequently, prompt-based learning involves training a language model on the converted prompt-based dataset. The essence of prompt-based learning is to modify the input into prompts and embed the anticipated output as unfilled blanks within the prompt.

To investigate the effects of prompt-based design in visual applications, models like CLIP \cite{Radford2021Learning} and ALIGN \cite{Jia2021Scaling} have achieved remarkable performance on image recognition tasks. They formulate the objectives as descriptive texts and transit the classification problem into video-text matching. Besides, ActionCLIP \cite{Wang2021Actionclip} proposes a prompt-based paradigm specific to action recognition tasks, aiming to recognize single actions in short video clips. These methods demonstrate the potential of prompt-based learning to motivate the development of visual tasks with multi-modal feature representation.

To this end, we attempt to rethink the effectiveness of frame-wise feature extraction in the TAS task, for which previous action segmentation methods uniformly utilize the pre-trained I3D \cite{Carreira2017Quo} model. Although the I3D-based features maintain advantageous capacity due to integrating information from the RGB stream and optical-flow stream, they might not be sufficient to construct representative embeddings experiencing complex scenarios, particularly for CPR actions. One closely related work \cite{Li2022Bridge} that arose recently, called Br-Prompt, is of significant importance for instructing the feature extraction model equipped with a prompt-based paradigm. On top of that, our approach extends more details elaborated for effective feature extraction of CPR actions in favor of downstream tasks.

\begin{figure*}[htb]
\centerline{\includegraphics{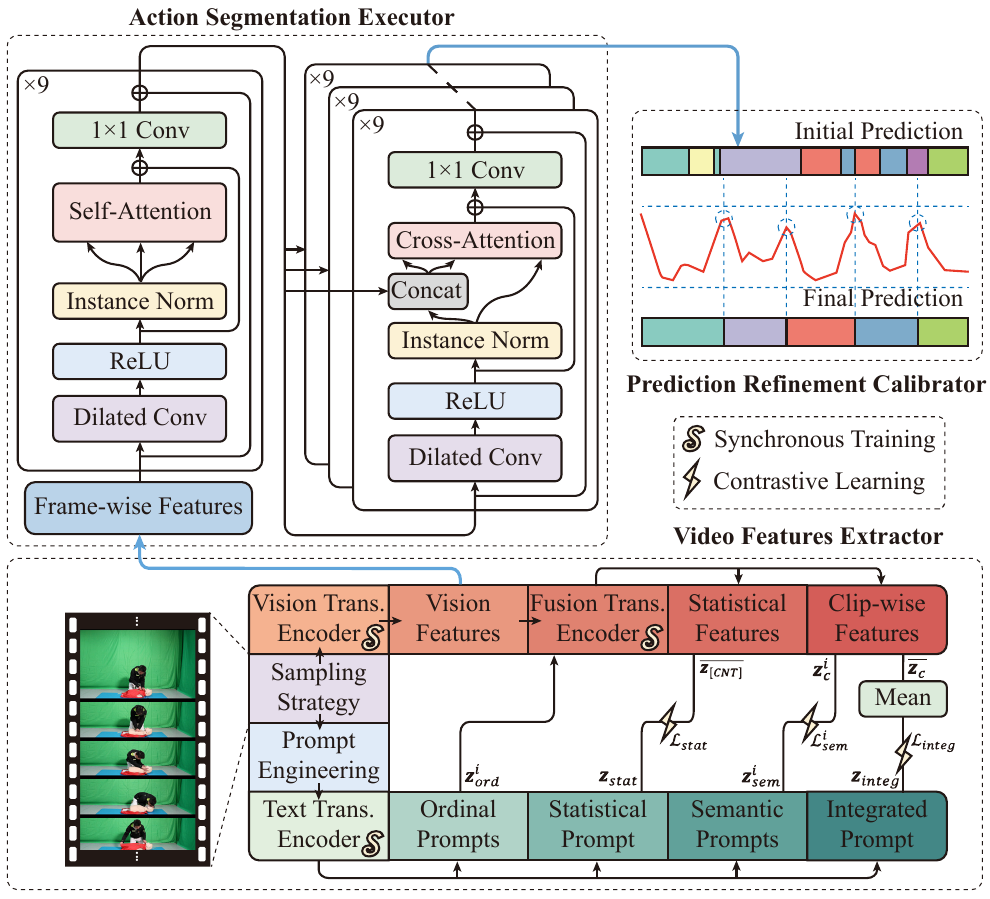}}
\caption{Overview of PhiTrans pipeline for cardiopulmonary resuscitation action segmentation. The VFE module at the bottom plays a role in generating frame-wise features, which will serve as the input to the top left ASE module to achieve the initial predictions. The PRC module in the top right further implements calibration to yield the eventual predictions. $\mathcal{L}_{stat}$, $\mathcal{L}_{sem}^{i}$, and $\mathcal{L}_{integ}$ in the VFE module are three elaborate contrastive learning losses. Vision Trans. Encoder, Text Trans. Encoder, and Fusion Trans. Module are synchronously trained. Diverse colors in the PRC module represent action categories, and their length indicates the duration of the corresponding action. Best viewed in color.}
\label{fig:phiTrans_model}
\end{figure*}

\section{Methods}
In this section, we introduce the proposed Prompt-enhanced hierarchical Transformer (PhiTrans) with clear motivation and objective. Then we distill its integral modules.

\subsection{Overall Pipeline}
As declared in Section~\ref{sec:introduction}, our motivation is to investigate \emph{how better could we elevate CPR instruction with action segmentation}, which nails down our objective of devising an exact architecture for CPR action segmentation to excavate its positive effects on CPR instruction.

The overall pipeline of our approach is illustrated in Fig.~\ref{fig:phiTrans_model}. We first apply Video Features Extractor (VFE) to extract frame-wise features representing potentially involved CPR actions and transition semantics. After that, Action Segmentation Executor (ASE) generates an initial prediction for the input video, explicitly identifying ordinal CPR actions. Finally, to effectively alleviate over-segmentation issues that might give rise to the misclassification of CPR actions, we adopt Prediction Refinement Calibrator (PRC) to refine the performance, thereby producing a final assessment enclosing less boundary ambiguity.

We argue that the aforementioned modules are indispensable, separately playing distinct roles and ultimately bringing down-to-earth performance to the intact model. To support our claims, more module details and corresponding impacts are described below.

\subsection{Video Features Extractor}
The Video Features Extractor (VFE) module is responsible for extracting discriminative frame-wise features of CPR videos, which subsequently become the input of the action segmentation module. As demonstrated at the bottom of Fig.~\ref{fig:phiTrans_model}, raw video frames are primarily sampled into neat clips with fixed lengths for local recognition and training efficiency. On top of retaining pre-established semantics by prompt engineering, the VFE module implements visual-linguistic contrastive learning to fuse multi-modal knowledge. Ternary Transformer encoders unify the framework and the synchronous training finally empowers the vision Transformer encoder to produce representative frame-wise features for input videos.

\subsubsection{Sampling strategy}
To unify the inputs and mitigate the footprint burden, we implement a sampling strategy for the raw CPR videos to generate a series of video clips. Specifically, a frame-wise sliding window approach is carried out with the downsampling rate ($ds$) of frames in each window and the overlapping rate ($ol$) between windows. The configuration of $ds$ and $ol$ will be illuminated in Section~\ref{sec:sampling strategy}. The sampling strategy allows the VFE module to concentrate on local dependencies with a fixed video length. Each generated video clip involves one or multiple CPR actions, leading to diverse receptive fields to feature extraction with training efficiency. Moreover, this strategy is vital for augmenting data and empowers the robustness of the VFE module.

\subsubsection{Prompt engineering}
In order to acquire informative semantic merits for a CPR video clip, prompt engineering resorts to the idea of embedding the expected output string into the input template in a cloze test-like form. We simulate the implementation inspired by Br-Prompt that is non-discriminatorily applied to TAS tasks. In particular, we establish 15 advantageous semantics for the CPR actions appeared in the custom dataset. Therefore, four types of prompts are available to record properties of CPR behaviors such as location, quantity, semantics, and integrality. More specifically, The ordinal prompt $\boldsymbol{z}_{ord}^{i}$ adopts the format as “\emph{this is the} \{\underline{$\mathrm{i^{th}}$}\} \emph{action in the video}” to simply captures the position of each state. The statistical prompt $\boldsymbol{z}_{stat}$ counts the number of CPR actions of a sequence with a neat format as “\emph{this video clip contains} \{\underline{number of CPR behaviors}\} \emph{actions in total}”. To investigate pronounced semantic analysis, we utilize the format “\{\underline{$\mathrm{i^{th}}$}\}\emph{, the person is performing the action step of} \{\underline{a certain CPR behavior}\}” as the semantic prompt $\boldsymbol{z}_{sem}^{i}$ focusing on adjacence and interspersion of actions. Eventually, we regard aggregated semantic prompts in a CPR video clip as the integrated prompt $\boldsymbol{z}_{integ}$, representing global semantic information. The motivation of prompt engineering aims to learn inter- and intra- affinities of behaviors within CPR videos.

\subsubsection{Visual-linguistic contrastive learning}
Given massive prompts by the prompt engineering, the VFE module ponders fabulous representation of the CPR video clips through visual-linguistic contrastive learning. To be legible, video clip $c$ and its text description $t$ are introduced to a visual encoder and a text encoder to obtain the corresponding representation $\boldsymbol{z}_c$ and $\boldsymbol{z}_t$, respectively, and the cosine similarity between the two is expressed as:
\begin{equation}
s(\boldsymbol{z}_c,\boldsymbol{z}_t)=\frac{\boldsymbol{z}_c\cdot \boldsymbol{z}_t}{\vert \boldsymbol{z}_c\vert \vert \boldsymbol{z}_t\vert}
\end{equation}
The batch similarity matrix $S$ for the video clip feature $\mathcal{Z}_c$ and the text feature $\mathcal{Z}_t$ with batch size $B$ is:
\begin{equation}
S(\mathcal{Z}_c,\mathcal{Z}_t)=\begin{bmatrix} s(\boldsymbol{z}_{c_1}, \boldsymbol{z}_{t_1}) &\cdots &s(\boldsymbol{z}_{c_1}, \boldsymbol{z}_{t_B}) \\
\vdots &\ddots &\vdots \\
s(\boldsymbol{z}_{c_B}, \boldsymbol{z}_{t_1}) &\cdots &s(\boldsymbol{z}_{c_B}, \boldsymbol{z}_{t_B})
\end{bmatrix}
\end{equation}

We define the ground-truth batch similarity matrix $GT$, where the similarity of the correct pair is set to 1, and contrastively the error pair is set to 0. The objective of learning is to maximize the similarity between $S$ and $GT$.

Here we adopt the KL divergence (Kullback–Leibler divergence) as the contrastive loss. For instance, $N\times N$ matrices $P$ and $Q$, its brief form is defined:
\begin{equation}
D_{KL}(P\Vert Q)=\frac{1}{N^{2}}\sum_{i=1}^{N}\sum_{j=1}^{N}P_{ij}\log \frac{P_{ij}}{Q_{ij}}
\end{equation}

In this way, given dual modal similarity matrix $S_{C}$ and $S_{T}$, we leverage the visual-linguistic contrastive loss as:
\begin{equation}
\mathcal{L}=\frac{1}{2}[D_{KL}(S_{C}\Vert GT)+D_{KL}(S_{T}\Vert GT)]
\end{equation}

To explain comprehensive details while avoiding ambiguous perplexity, we first employ a semantic loss $\mathcal{L}_{sem}^{i}$, in which visual features $\boldsymbol{z}_{c}^{i}$ involving ordinal contents of the video clip contend against commensurate textual semantic representation $\boldsymbol{z}_{sem}^{i}$, thereby maintaining substantial semantic concernments. Meanwhile, the average pooling video clip features $\overline{\boldsymbol{z}_{c}}$ containing ordinal collections compete with integrated prompts representation $\boldsymbol{z}_{integ}$ via the integrated loss $\mathcal{L}_{integ}$ to allow holistic demonstration. Furthermore, to transparently exert a profound influence on the quantitive information, a statistical loss $\mathcal{L}_{stat}$ is applied by way of contrasting average pooling enumeration features $\overline{\boldsymbol{z}_{[CNT]}}$ with statistical prompts representation $\boldsymbol{z}_{stat}$. We ablate these loss components in Section~\ref{sec:loss ablation}.

Finally, the united loss of the model is manifested as:
\begin{equation}
\mathcal{L}=\sum_{i=1}^{K} \mathcal{L}_{sem}^{i}+\mathcal{L}_{integ}+\mathcal{L}_{stat}
\end{equation}

\subsection{Action Segmentation Executor}
The Action Segmentation Executor (ASE) module takes frame-wise features derived from the VFE module as input and produces moderate CPR action segmentation predictions. The structure of the ASE module is illustrated in the upper left of Fig.~\ref{fig:phiTrans_model}. Generally, the ensemble is a hierarchical encoder-decoder architecture, wherein the output of the encoder as initial predictions will flow in ternary decoders for distillation. At the same time, the ASE module attempts to acquire hierarchical perception with window-perceptive self-attention and dilated temporal convolution throughout the framework.

\subsubsection{Hierarchical encoder-decoder}
ASE module is composed of one encoder and three identical decoders. Each of them contains nine blocks to provide hierarchical representation. After generating the initial prediction by the encoder, three decoders perform a refinement process to boost the performance. The input to the encoder is a sequence of pre-extracted frame-wise features in sizes $D\times S$, where $D$ refers to the feature dimension of each frame and $S$ represents the total number of frames of the input video. The first layer of the encoder uses a fully connected layer to adjust the input feature dimension. Subsequently, each encoder block utilizes a dilated temporal convolution as a feed-forward layer, which is followed by a ReLU activation function and instance normalization, connecting the single-head self-attention layer. A residual concatenation is taken between the two layers, and then the output dimension is reshaped by 1 × 1 convolution to join the next encoder block. The final encoder block outputs the initial prediction $p_{e}\in R^{S\times C}$ by passing a fully connected layer, where C represents the number of action categories. Afterward, the decoder takes the initial prediction as input and is arranged in accordance with the encoder as a whole, except for the cross-attention layer. The cross-attention layer combines the results from the encoder with the output from the previous layer, treating the aggregation as the query $Q$ and the key $K$, and the output from the previous layer as the value $V$. The advantage of this manner is that the frame-wise confidence scores from the encoder can be involved in the refinement stage by generating attention weights, and these attention weights are utilized for linear concatenation, without affecting the feature space $V$ itself. In the end, the model makes the best use of three identical decoders to hierarchically produce the final prediction.

\subsubsection{Hierarchical perception}
Two hierarchical strategies are simultaneously adopted to scratch multi-scale receptive fields by enlarging the window size of the self-attention layer and dilated temporal convolution throughout the ASE module. Considering videos in the CPR dataset tend to cover thousands of frames, it is fairly demanding to seize significant vision plots for the self-attention layer within each block of the encoder or the decoders. We follow the spirit of hierarchical representation patterns proposed in \cite{Yi2021Asformer} to mitigate this issue. Concretely, such a strategy first concentrates on the local semantics and then gradually enlarges the receptive field to acquire the global concepts, allowing the model to learn extrinsic-to-intrinsic knowledge of CPR actions displayed in the videos. In addition, both apparent differences and subtle discrepancies are taken into account, yielding a more specific fashion to the traits of CPR actions. Practically, the window-perceptive self-attention layer calculates attention weights with each particular frame within its local window at a $w$ scale, which is doubled as the blocks stack deeper (i.e., $w=2^i,\ i=1,2,\ldots$). Similarly, to introduce constructive local inductive bias, we follow the dilated temporal convolution as utilized in the TAS task by expanding the kernel size consistent with the self-attention layer.

\subsection{Prediction Refinement Calibrator}
The Prediction Refinement Calibrator (PRC) module intends to remarkably alleviate over-segmentation errors of the predictions from the ASE module. The PRC module allies with and refines the ASE module by duplicating the segmentation pipeline but replacing the objective from action predictions with boundary probabilities. In particular, a comprehensive loss function serves as the motivation for jointly training the ASE and PRC modules. The overall loss function is a combination of the action segmentation loss and the boundary regression loss, with correlative ablation conducted in Section~\ref{sec:loss ablation}. As shown in the upper right of Fig. 1, the generated boundary probability curves further calibrate the action predictions from the ASE module with less over-segmentation issues and present the final CPR action segmentation predictions.

\subsubsection{Boundary probability calibration}
To clarify the assets of the PRC module, we depict its intrinsic concept for boundary probability calibration. During the inference period, the PRC module regresses frame-wise action boundary possibilities $P \in [0,1]^{S}$. Subsequently, the action boundaries $B \in \{0,1\}^{S}$ are determined by electing multiple $P_{s}$ from $P$ as the local maximum and simultaneously fulfilling the exceeding condition of the threshold $p=0.5$.

In practice, the ASE module primarily generates initial predictions by assigning the action categories to the action segments with potential over-segmentation issues. Then the calculated action boundaries $B$ compartmentalize the video clip into refined action segments, each of which contains only one action both theoretically and practically. The retained action categories by majority voting for each action segment reach the final prediction of the entire model.

\subsubsection{Action segmentation loss}
The overall action segmentation loss is defined as:
\begin{equation}
\mathcal{L}_{as}=\frac{1}{G}\sum_{g}\big(\mathcal{L}_{cls}^{*}+\mathcal{L}_{smo}\big)
\end{equation}

More specifically, the classification loss $\mathcal{L}_{cls}^\ast$ utilizes median frequency balancing, where the action weights of each action category in the temporal action segmentation task are calculated by dividing the mean frequency of each action category by the frequency of each action class. Concerning the smoothing loss $\mathcal{L}_{smo}$, we implement Gaussian Similarity-weighted Truncated Mean Squared Error (GS-TMSE), which penalizes all frames in a video clip for the purpose of smoothing the transition of action probabilities between frames while preventing them from interfering with the frames where actions de facto transition. Its concrete form is as follows:
\begin{gather}
\mathcal{L}_{GS-TMSE}=\frac{1}{SC}\sum_{s,c}\exp \Big(-\frac{\Vert \boldsymbol{x}_{s}-\boldsymbol{x}_{s-1}\Vert^{2}}{2\sigma^{2}}\Big)\delta_{s,c}^{2} \\
\delta_{s,c}=\min \{\vert \log p_{s,c}-\log p_{s-1,c}\vert ,\ \tau\}
\end{gather}

where $\boldsymbol{x}_{s}$ is the similarity index of the frame $s$, $\sigma$ denotes the variance and is simply set to 1, and the threshold $\tau$ is set to 4. The advantageous implementation benefits from the property of GS-TMSE. For brevity, the Gaussian kernel based on the similarity of frames (frame-wise features in our experiments), the function punishes contiguous but discriminative frames with merely a small weight. We average the losses of $G$ stages (in our framework $G=4$) as the overall action segmentation loss.

\subsubsection{Boundary regression loss}
We employ a weighted logistic regression loss function to endow our model with boundary-aware capacity, which is defined as:
\begin{equation}
\mathcal{L}_{br}=\frac{1}{SG}\sum_{g}\sum_{s=1}^{S}\big(w_{p}y_{s}\cdot \log p_{s}+(1-y_{s})\cdot \log (1-p_{s})\big)
\end{equation}

where $y_s$ and $p_s$ represent the ground-truth action boundary labels and predicted action boundary possibilities for the frame $s$, respectively. Since the number of action boundary frames is much smaller than that of other frames, the factor $w_p$ is kindly devised to samples with weight positive. To further clarify $w_p$, it comes from the reciprocal ratio of positive data points over the entire training data.

\begin{table}[htb]
\caption{Summary of datasets for model development and application}
\label{tab:dataset summary}
\setlength{\tabcolsep}{3pt}
\centering
\begin{tabular}{c|cccccc}
\hline
    Dataset & \emph{videos} & \emph{classes} & \emph{duration} & \emph{fps} & \emph{instances} & \emph{view} \\ \hline
    50Salads & 50 & 17 & 6.4 & 30 & 20 & top \\ 
    Breakfast & 1,712 & 48 & $\approx$1 & 15 & 6 & third person \\ 
    GTEA & 28 & 11 & $\approx$1 & 15 & 20 & egocentric \\ \hline
    \textbf{CPR} & \textbf{99} & \textbf{15} & \textbf{2} & \textbf{15} & \textbf{17} & \textbf{third person} \\ \hline
    \end{tabular}
\end{table}

\begin{figure*}[htb]
\centering
\includegraphics{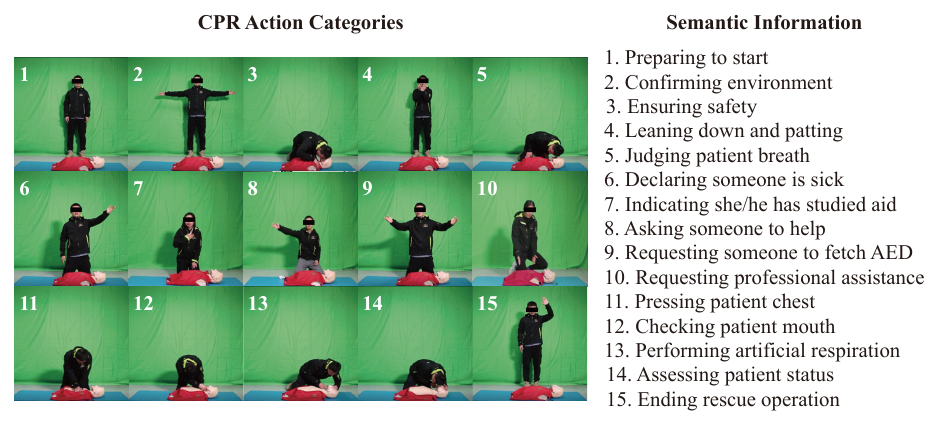}
\caption{Visualization of 15 action categories recorded in the CPR dataset and their corresponding semantic information. Best viewed in color.}
\label{fig:PhiTrans_datasets}
\end{figure*}

\section{Dataset}
We primarily compare the performance of various models on three challenging public datasets that are widely recognized in the TAS task concentrating on catering behaviors, whereby the optimal architecture will be established especially for CPR action segmentation. We argue for the reasonability of selecting these three datasets on account of analogous representation patterns compared to the CPR actions, wherein the behaviors are globally invariant and locally distinct. More importantly, we introduce the preprocesses of the custom CPR dataset and explain its implication. The essential takeaway information for these four datasets is summarized in Table~\ref{tab:dataset summary}.

\subsection{Preliminary Public Datasets}
50Salads dataset \cite{stein2013combining} is beneficial for research in action recognition, activity detection, process tracking, etc., which collects 50 videos from 25 participants preparing two different kinds of mixed salads, and contains more than 4 hours of video data. Each video lasts 6.4 minutes long on average and consists of about 20 action instances. All videos are recorded from a top-down view, including 17 action category annotations.

Breakfast dataset \cite{kuehne2014language} is related to the preparation of daily breakfast. It comprises 18 different kitchen scenes with diverse backgrounds conducive to monitoring and analyzing daily human activities. The dataset contains 1,712 videos with 48 action annotations recorded from a third-person view of 52 participants, wherein each video represents an average of 6 action instances.

GTEA dataset \cite{fathi2011learning} contains 28 egocentric videos with 11 action categories involving daily kitchen activities performed by 4 participants. On average, each video has 20 action instances and is about a half-minute long.

\subsection{Custom CPR Dataset}
We collect the CPR dataset with ethical approval and notification of subjects with the assistance of MIRAGESTARS Inc., which involves 99 videos of participants performing the whole process of cardiopulmonary resuscitation in a standard green screen laboratory environment. This work is conducted in adherence to the tenets of the Declaration of Helsinki, and ethics approval is obtained by the Ethics Committee of the Shenzhen International Graduate School of Tsinghua University Submission F111/2022.

Each video has about two minutes duration on average, containing exactly 15 action categories shown in Fig.~\ref{fig:PhiTrans_datasets} with their representative frames and the corresponding semantic information. The raw videos shot are identically transformed utilizing the stream processing tool called FFmpeg. More concretely, the frame rate and resolution of videos are decreased from 25 fps, 2k to 15 fps, 720p respectively, on account of alleviating the calculating burden of proposed models, while ensuring the accessibility of the pipeline and the efficiency of the architecture to the maximum extent. It is worth noting that the simplifications above significantly reduce the video size from 320 MB -- 1.1 GB to 1.9 MB -- 5.3 MB, followed by wiping out the audio channel. These transformations sufficiently consider that CPR action segmentation generally depends on integral gestures rather than pixel-wise identification, whereby hardly any practical loss will be generated. After that, these videos are labeled at a frame-level by related trainers. The crafted CPR dataset possesses diverse challenges, including but not limited to the transience of partial actions, resemblance among actions, and out-of-place actions.

The motivation of this dataset is prone to assist CPR instruction by automatically identifying potential omission, repetition, or out-of-order situations at a second-wise level, even if taking frame-wise misregistration into consideration. Furthermore, we adopt to implement four-fold nested cross validation to minimize latent optimistically biased evaluation, thus fairly revealing the performance of the presented model.

\section{Experiments and Results}
\label{sec:experiments and results}
\subsection{Evaluation Metrics}
\subsubsection{Frame-wise accuracy (Acc.)}
Frame-wise accuracy is commonly used as an evaluation metric for action segmentation \cite{stein2013combining,kuehne2014language,Lea2016Segmental}, whereas it is easily affected by long-duration actions and not sensitive to over-segmentation issues.

\begin{table*}[htb]
\caption{Action segmentation performance of various state-of-the-art (SOTA) models on GTEA, 50Salads datasets and Breakfast}
\label{tab:model selection}
\resizebox{\linewidth}{!}{
\begin{tabular}{c|ccc|c|c|ccc|c|c|ccc|c|c}
\hline
    \textbf{Dataset} & \multicolumn {5}{c|}{\textbf{GTEA}} & \multicolumn {5}{c|}{\textbf{50Salads}} & \multicolumn {5}{c}{\textbf{Breakfast}} \\ \hline
    \textbf{Model} & \multicolumn{3}{c|}{\emph{F1@\{10,25,50\}}} & \emph{Edit} & \emph{Acc.} & \multicolumn{3}{c|}{\emph{F1@\{10,25,50\}}} & \emph{Edit} & \emph{Acc.} & \multicolumn{3}{c|}{\emph{F1@\{10,25,50\}}} & \emph{Edit} & \emph{Acc.} \\ \hline
    MS-TCN \cite{Farha2019Ms} & 85.8 & 83.4 & 69.8 & 79.0 & 76.3 & 76.3 & 74.0 & 64.5 & 67.9 & 80.7 & 52.6 & 48.1 & 37.9 & 61.7 & 66.3 \\ 
    DTGRM \cite{Wang2021Temporal} & 87.8 & 86.6 & 72.9 & 83.0 & 77.6 & 79.1 & 75.9 & 66.1 & 72.0 & 80.0 & 68.7 & 61.9 & 46.6 & 68.9 & 68.3 \\ 
    BCN \cite{Wang2020Boundary} & 88.5 & 87.1 & 77.3 & 84.4 & \underline{79.8} & 82.3 & 81.3 & 74.0 & 74.3 & 84.4 & 68.7 & 65.5 & 55.0 & 66.2 & \underline{70.4} \\ 
    MS-TCN++ \cite{Li2020MS} & 88.8 & 85.7 & 76.0 & 83.5 & 80.1 & 80.7 & 78.5 & 70.1 & 74.3 & 83.7 & 64.1 & 58.6 & 45.9 & 65.6 & 67.6 \\ 
    ASRF \cite{Ishikawa2021Alleviating} & 89.4 & 87.8 & \underline{79.8} & 83.7 & 77.3 & 84.9 & \underline{83.5} & \underline{77.3} & 79.3 & 84.5 & 74.3 & 68.9 & \underline{56.1} & 72.4 & 67.6 \\ 
    SSTDA \cite{Chen2020Action} & 90.0 & \underline{89.1} & 78.0 & 86.2 & \underline{79.8} & 83.0 & 81.5 & 73.8 & 75.8 & 83.2 & \underline{75.0} & \underline{69.1} & 55.2 & \underline{73.7} & 70.2 \\ 
    \textbf{ASFormer} \cite{Yi2021Asformer} & \underline{90.1} & 88.8 & 79.2 & \underline{84.6} & 79.7 & \underline{85.1} & 83.4 & 76.0 & \underline{79.6} & \underline{85.6} & \textbf{76.0} & \textbf{70.6} & \textbf{57.4} & \textbf{75.0} & \textbf{73.5} \\ 
    \textbf{Br-Prompt} \cite{Li2022Bridge}+ASFormer & \textbf{94.1} & \textbf{92.0} & \textbf{83.0} & \textbf{91.6} & \textbf{81.2} & \textbf{89.2} & \textbf{87.8} & \textbf{81.3} & \textbf{83.8} & \textbf{88.1} & N/A & N/A & N/A & N/A & N/A \\ \hline
    \end{tabular}
    }
\end{table*}

\subsubsection{Segmental edit score (Edit)}
Segmental edit score \cite{Lea2016Learning} is used to assess the model performance in predicting the ordering of action segmentation without being affected by minor temporal shifts. Once proposed, the segmental edit score has been widely used in many temporal action segmentation tasks \cite{Wang2020Boundary,Wang2021Temporal,Huang2020Improving,Chen2020Action,Lea2016Segmental} since it combines the assessment of accuracy and efficiency into a single metric. There is considerable uncertainty about when one action will cease and another will begin. Typically, in practical applications such as surgical workflow assessment, the accurate temporal continuity of surgical operations tends to be more crucial than precise temporal segmentation, as the same goes for CPR instruction.

\subsubsection{Segmental F1 score with overlapping threshold k (F1@k)}
Segmental overlap \emph{F1} score \cite{Lea2017Temporal} has three distinctive characteristics: 1) penalizes over-segmentation errors; 2) ignores minor temporal shifts between the predictions and ground truth; 3) is determined by the total number of actions but does not depend on the duration of each action instance. By comparing the Intersection over Union (IoU) score of the predictions and ground truth, if the threshold $\tau=\frac{k}{100}$ is exceeded, it is determined as true positive, otherwise as true negative, where $k=10,25,50$ are adopted in temporal action segmentation tasks. The $precision$ and $recall$ are defined as $precision=\frac{true\ positives}{true\ positives+false\ positives}$, $recall=\frac{true\ positives}{true\ positives+false\ negatives}$, respectively. Then \emph{F1@k} value can be computed from $F1=2\times \frac{precision*recall}{precision+recall}$.

\subsection{Module Selection for Action Segmentation}
It is non-trivial to select an appropriate and distinctive backbone to approach our goal of CPR action segmentation. Table~\ref{tab:model selection} reveals the best temporal action segmentation models on GTEA, 50Salads, and Breakfast datasets. Comparing the performance on public datasets is intuitively convincing, for which we gaze at the optimal one, ASFormer \cite{Yi2021Asformer}. Particularly, we overlook Br-Prompt \cite{Li2022Bridge} in this subsection, which plays a role in feature representation, which enhances the action segmentation models rather than serves as one of them.

Considering the property of the collected CPR dataset, we lay more emphasis on the analysis capacity to confront more enormous and complex data. Concretely, on the GTEA and 50Salads datasets with slightly smaller video volumes, we find ASFormer exhibits inference accuracies comparable to ASRF \cite{Ishikawa2021Alleviating} and SSTDA \cite{Chen2020Action}. While carried out on the Breakfast dataset with a multitude of videos, the model explicitly yields state-of-the-art performance. Through observing the performance on three public datasets, ASFormer illustrates the application of Transformer in the TAS task, affirming the non-negligible capacity of temporal representation and persistent sensitivity of long-term relationships. Due to the convenient scalability and robustness of ASFormer, we adopt this model as a vanilla backbone for CPR action segmentation.

\begin{figure*}[htb]
\centerline{\includegraphics{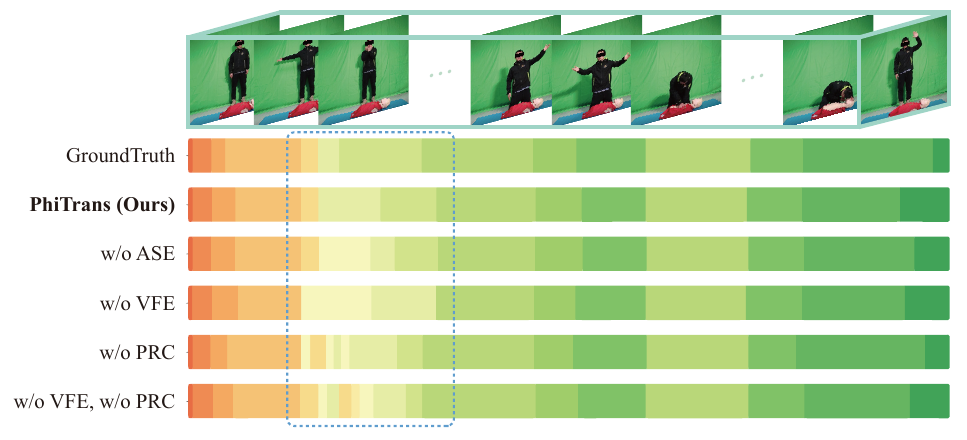}}
\caption{Visualization of PhiTrans and related variants for temporal action segmentation on the CPR dataset. Diverse colors represent action categories, and their length indicates the duration of the corresponding action. Best viewed in color.}
\label{fig:PhiTrans_results}
\end{figure*}

\begin{table}[htb]
\caption{Action segmentation performance with features following various window configurations on the CPR dataset (\#split 1)}
\label{tab:sampling strategy}
\setlength{\tabcolsep}{5pt}
\centering
\begin{tabular}{c|ccc|c|c}
\hline
     \textbf{CPR} & \multicolumn{3}{c|}{\emph{F1@\{10,25,50\}}}& \emph{Edit} & \emph{Acc.} \\ \hline
    Baseline & 95.8 & 93.6 & 89.8 & 94.2 & 89.5 \\ 
    \emph{ds}=[2, 4] \& \emph{ol}=[2, 2] & 95.3 & 94.8 & 91.5 & 93.3 & \textbf{92.0} \\ 
    \emph{ds}=[2, 4, 8] \& \emph{ol}=[4, 2, 1] & \textbf{96.5} & 95.5 & 92.0 & 94.8 & 91.0 \\
    \emph{ds}=[4, 8, 12] \& \emph{ol}=[2, 1, 1] & 96.0 & \textbf{95.7} & \textbf{92.7} & \textbf{95.0} & 91.8 \\ \hline
    \end{tabular}
\end{table}

\subsection{Sampling Strategy Selection for Feature Extraction}
\label{sec:sampling strategy}
In the process of extracting video frame-wise features, we adopt a specific sampling strategy dependent on pertinent experiments to generate video clips of a fixed length, which serve as the input of the video features extractor. In general, each video clip adopts a 16-frame window. We are mainly concerned with the downsampling rate (\emph{ds}) of frames in each window, and the overlapping rate (\emph{ol}) between two windows. It should be emphasized that both the downsampling rate and the overlapping rate are heuristically selected according to the characteristics of the dataset. Concretely, longer windows lead to sparse information about each action, while shorter ones are even unable to contain more than two actions. We empirically take both long and short windows into consideration, which produce excellent feature extraction performance by involving multi-scale information.

Diverse window configurations are displayed in Table~\ref{tab:sampling strategy}, which indicate the non-trivial importance of the downsampling rate and the overlapping rate. The collective results are retained on the custom CPR dataset (split \#1) under the support of the ASE module as the action segmentation backbone. Multiple downsampling rates and overlapping rates are adopted simultaneously as mentioned above. In addition, training efficiency is an essential factor to be carefully considered. Complicated windows conduce to substantial training time, while simple ones may fail to grab discriminative information. In the light of this, we conduct three groups of window arrangements to seek out the optimal one. Here we adopt the preceding Transformer-based method \cite{Yi2021Asformer} as the vanilla baseline.

\begin{table*}[htb]
\caption{Model performance and module ablation of PhiTrans on the CPR dataset}
\label{tab:module ablation}
\setlength{\tabcolsep}{5pt}
\centering
\begin{tabular}{c|ccc|ccc|c|c}
\hline
\multirow{2}{*}{Model with Variants} & \multicolumn{3}{c|}{Components}                                  & \multicolumn{3}{c|}{\multirow{2}{*}{\emph{F1@\{10,25,50\}}}} & \multirow{2}{*}{\emph{Edit}} & \multirow{2}{*}{\emph{Acc.}} \\ \cline{2-4}
                          & \multicolumn{1}{c|}{ASE} & \multicolumn{1}{c|}{VFE} & PRC        & \multicolumn{3}{c|}{}                                        &                              &                              \\ \hline
\textbf{PhiTrans (Ours)}  & \checkmark               & \checkmark               & \checkmark & \textbf{95.8}      & \textbf{95.3}      & \textbf{91.7}      & \textbf{94.6}                & \textbf{91.1}                \\ \hline
w/o ASE                   & $\times$                 & \checkmark               & \checkmark & 94.3               & 93.0               & 88.5               & 92.9                         & 88.6                         \\
w/o VFE                   & \checkmark               & $\times$                 & \checkmark & 94.1               & 91.7               & 87.1               & 91.8                         & 87.6                         \\
w/o PRC                   & \checkmark               & \checkmark               & $\times$   & 95.3               & 94.5               & 91.5               & 93.6                         & 91.1                         \\ \hline
w/o VFE, w/o PRC          & \checkmark               & $\times$                 & $\times$   & 95.3               & 94.2               & 89.5               & 93.4                         & 89.7                         \\ \hline
\end{tabular}
\end{table*}

When the downsampling rates are 2, 4, and correspondingly the overlapping rates are 2, 2, redundant local features are captured. Each window contains only few actions, where a good deal of frames belong to the same action, resulting in poor effects. Similarly, employing 2, 4, and 8 downsampling rates and 4, 2, and 1 overlapping rates though reaches a high \emph{F1@50} as 96.5, its overlapping rate of 4 misleads the module extensively reusing the same information, decreasing the abundance of video features. We finally adopt downsampling rates as 4, 8, and 12 corresponding to the overlapping rates of 2, 1, and 1, which balance the expressions derived from various receptive fields with rational training efficiency.

\subsection{Model Performance and Ablation Study of PhiTrans}
In this subsection, we first present the favorable performance of PhiTrans both on the custom and public datasets. Then module ablations are demonstrated both quantitatively and qualitatively. Finally, we verify the effect of every loss component implemented in the proposed model.

\subsubsection{Model Performance of PhiTrans}
Our model accomplishes the objective to serve as a productive toolbox assisting CPR instruction with action segmentation, with the intact model performance manifested in Fig.~\ref{fig:PhiTrans_results} and Table~\ref{tab:module ablation}. On the whole, our model is adequate to approach the challenge of assisting CPR instruction, performing well on all metrics surpassing 91.0\%. The comprehensive performance reveals the effectiveness and robustness of PhiTrans. More importantly, PhiTrans reaches 94.6\% on \emph{Edit}, a metric which reflects the model ordinal predicting performance wiping out fine-drawn temporal shifts. That means our model is prone to understand the chronological relationship of CPR actions that practically necessitates. In addition, 95.8\%, 95.3\%, and 91.7\% on \emph{F1@\{10,25,50\}} indicate that PhiTrans successfully penalizes over-segmentation errors highlighted in the TAS task. Though not designed to pursue a state-of-the-art performance on the public dataset, PhiTrans somewhat outperforms cutting-edge models, revealed in Table~\ref{tab:model performance}. It gains +0.8\% improvement on \emph{F1@50}, with other metrics comparable to previous state-of-the-art, which further manifests the effectiveness of PhiTrans on the general TAS task.

\begin{figure}[htb]
\centerline{\includegraphics[width=\linewidth]{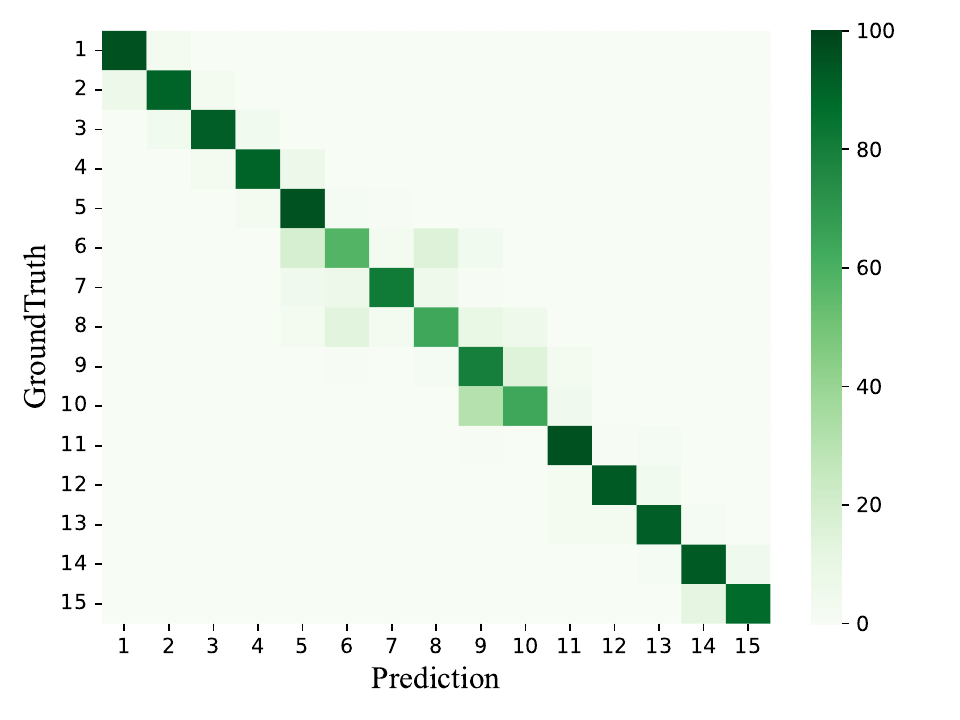}}
\caption{Confusion matrix for the custom CPR dataset. The numbers on the horizontal and vertical axes represent the ordinal CPR actions. The numerical unit in the confusion matrix is expressed in percentages. Best viewed in color.}
\label{fig:confusion_matrix}
\end{figure}

Moreover, it is possible to explain the segmentation errors that the model induces through empirical observation and analysis. Fig.~\ref{fig:confusion_matrix} visualizes the confusion matrix of the CPR dataset, which witnesses dominant segmentation performance on the majority of CPR actions by PhiTrans. Although our model exhibits hesitation in distinguishing actions ranging from 6 (\emph{declaring someone is sick}) to 10 (\emph{requesting professional assistance}), it is rational since these actions primarily differ in terms of vocal expression by the subjects, which is not incorporated into our modality for accommodating complex real-world scenarios.

\subsubsection{Module Ablation of PhiTrans}
Table~\ref{tab:module ablation} quantitatively reveals the effect of each module on holistic model performance. We detach each part independently to observe any drop. We experiment w/o ASE by replacing the ASE module with the approved backbone \cite{Farha2019Ms}, w/o VFE by adopting ResNet152 backbone-based frame-wise features as input, and w/o PRC by merely applying loss functions proposed in previous methods \cite{Yi2021Asformer}. The results show that lacking any module will lead to an apparent performance drop on all metrics, manifesting the essentiality of each module. Worst of all, the loss of the VFE module (w/o VFE) causes the model performance to reach a trough, leading to a disastrous result of the total -9.9\% on \emph{F1} and -2.8\%, -3.5\% on \emph{Edit} and \emph{Acc.}, respectively. Interestingly, the further removal of the PRC module on this basis (namely, w/o VFE, w/o PRC) instead presents a relative increase on all metrics. This situation can be explained by the fact that the PRC module tends to suppress over-segmentation issues excessively when the feature expression is insufficient. These results prove the indispensability of the VFE module for its splendid feature representation. Similarly, whichever module misses accounts for the loss of corresponding capacity.

Besides, to intuitively present effects of these modules, Fig.~\ref{fig:PhiTrans_results} visualizes the qualitative representation on the custom CPR dataset. Under the reinforcement of each module, PhiTrans could not only perceive the presence of delicate actions but alleviate over-segmentation issues to a great extent, as displayed in the blue dashed line. Three hands-on modules proportionally bring promising performance for the challenge of assisting CPR instruction.

\begin{table}[htb]
\caption{Model performance of PhiTrans on the 50Salads dataset}
\label{tab:model performance}
\setlength{\tabcolsep}{5pt}
\centering
\begin{tabular}{c|ccc|c|c}
\hline
    \textbf{50Salads} & \multicolumn{3}{c|}{\emph{F1@\{10,25,50\}}}& \emph{Edit} & \emph{Acc.} \\ \hline
    Cutting-edge \cite{Li2022Bridge} & 89.2 & 87.8 & 81.3 & \textbf{83.8} & 88.1 \\ 
    \textbf{PhiTrans (ours)} & \textbf{89.3} & \textbf{87.8} & \textbf{82.1} & 83.4 & \textbf{88.1} \\ \hline
    \end{tabular}
\end{table}

\begin{table}[htb]
\caption{Loss ablation of PhiTrans on the CPR dataset (\#split 1)}
\label{tab:loss ablation}
\setlength{\tabcolsep}{5pt}
\centering
\begin{tabular}{l|ccc|c|c}
\hline
    \textbf{VFE Loss components} & \multicolumn{3}{c|}{\emph{F1@\{10,25,50\}}}& \emph{Edit} & \emph{Acc.} \\ \hline
    $\mathcal{L}_{sem}$ & 94.1 & 91.6 & 88.1 & 92.7 & 88.7 \\ 
    $\mathcal{L}_{sem}+\mathcal{L}_{integ}$ & 95.1 & 93.9 & 90.1 & 92.8 & 90.0 \\ 
    $\mathcal{L}_{sem}+\mathcal{L}_{integ}+\mathcal{L}_{stat}$ & \textbf{96.2} & \textbf{95.5} & \textbf{91.4} & \textbf{95.5} & \textbf{90.5} \\ \hline\hline
    \textbf{PRC Loss components} & \multicolumn{3}{c|}{\emph{F1@\{10,25,50\}}}& \emph{Edit} & \emph{Acc.} \\ \hline
    $\mathcal{L}_{as}$ & 93.7 & 92.4 & 88.3 & 91.0 & 89.4 \\ 
    $\mathcal{L}_{as}+\mathcal{L}_{br}$ & \textbf{96.2} & \textbf{95.5} & \textbf{91.4} & \textbf{95.5} & \textbf{90.5} \\ \hline
    \end{tabular}
\end{table}

\subsubsection{Loss Ablation of PhiTrans}
\label{sec:loss ablation}
Table~\ref{tab:loss ablation} presents two types of loss components that arise in the proposed model, one for extracting frame-wise features and the other for refining segmentation. Specifically, all three VFE loss components contribute to capturing representative features. The semantic loss $\mathcal{L}_{sem}$ supervises distinctive features related to various CPR actions, laying the foundation for feature extraction. As the integrated $\mathcal{L}_{integ}$ and statistical loss $\mathcal{L}_{stat}$ are successively added, the model achieves respective improvement, indicating the indispensability of all losses. Additionally, it can be witnessed that both loss components utilized in the PRC module advocate the model performance by action segmentation control and boundary regression refinement. Notably, the boundary regression loss $\mathcal{L}_{br}$ brings +3.0\% on every \emph{F1} and \emph{Edit}), exactly showing its effectiveness to alleviate over-segmentation issues. Overall, these loss ablations certify the necessity and trait of losses implemented in our model.

\section{Conclusion and Discussion}
In this study, we originally deliberate a rewarding pipeline that assists the enhancement of CPR instruction via action segmentation through novel deep learning architectures. Specifically, we collect a custom CPR dataset involving videos of the whole process of cardiopulmonary resuscitation assessment along with corresponding frame-level semantic annotation. Accordingly, we devise a Prompt-enhanced hierarchical Transformer, called PhiTrans, especially for CPR action segmentation. PhiTrans consists of three integral modules: Video Features Extractor, Action Segmentation Executor, and Prediction Refinement Calibrator. Such an architecture adequately considers the characteristics of CPR actions, facilitating the capacity to capture transient CPR actions while maintaining impressive performance. Extensive ablation experiments present that PhiTrans inspires improvement compared to half-baked models on the CPR dataset.

In conclusion, it is non-trivial that PhiTrans is committed to CPR action segmentation, which is conducive to freeing experts of detecting and rectifying potential ordinal or oblivious mistakes made by subjects, thereby manifesting a compelling pipeline on elevating CPR instruction with action segmentation. The results of this research may serve as the cornerstone and offer a route towards developing a prospective orientation that would leverage fine-grained criteria like CPR action assessment.

\section{Acknowledgements}
This work was supported in part by the National Natural Science Foundation of China 31970752; Science, Technology, Innovation Commission of Shenzhen Municipality JCYJ20190809180003689, JSGG20200225150707332, JCYJ20220530143014032, KCXFZ20211020163813019, ZDSYS20200820165400003, WDZC20200820173710001, WDZC20200821150704001, JSGG20191129110812708.

\bibliographystyle{elsarticle-num}
\bibliography{PhiTrans}

\begin{thebibliography}{10}
\expandafter\ifx\csname url\endcsname\relax
  \def\url#1{\texttt{#1}}\fi
\expandafter\ifx\csname urlprefix\endcsname\relax\def\urlprefix{URL }\fi
\expandafter\ifx\csname href\endcsname\relax
  \def\href#1#2{#2} \def\path#1{#1}\fi

\bibitem{Brooks2022Optimizing}
S.~C. Brooks, G.~R. Clegg, J.~Bray, C.~D. Deakin, G.~D. Perkins, M.~Ringh,
  C.~M. Smith, M.~S. Link, R.~M. Merchant, J.~Pezo-Morales, et~al., Optimizing
  outcomes after out-of-hospital cardiac arrest with innovative approaches to
  public-access defibrillation: A scientific statement from the international
  liaison committee on resuscitation, Circulation 145~(13) (2022) e776--e801.

\bibitem{sandroni2021brain}
C.~Sandroni, T.~Cronberg, M.~Sekhon, Brain injury after cardiac arrest:
  pathophysiology, treatment, and prognosis, Intensive care medicine (2021)
  1--22.

\bibitem{Association1974Standards}
A.~H. Association, et~al., Standards for cardiopulmonary resuscitation (cpr)
  and emergency cardiac care (ecc), Jama 227 (1974) 837--860.

\bibitem{Lancet2018Out}
T.~Lancet, Out-of-hospital cardiac arrest: a unique medical emergency, The
  Lancet 391 (2018).

\bibitem{HasselqvistAx2015Early}
I.~Hasselqvist-Ax, G.~Riva, J.~Herlitz, M.~Rosenqvist, J.~Hollenberg,
  P.~Nordberg, M.~Ringh, M.~Jonsson, C.~Axelsson, J.~Lindqvist, et~al., Early
  cardiopulmonary resuscitation in out-of-hospital cardiac arrest, New England
  Journal of Medicine 372~(24) (2015) 2307--2315.

\bibitem{Wik1994Quality}
L.~Wik, P.~A. Steen, N.~G. Bircher, Quality of bystander cardiopulmonary
  resuscitation influences outcome after prehospital cardiac arrest,
  Resuscitation 28~(3) (1994) 195--203.

\bibitem{Gallagher1995Effectiveness}
E.~J. Gallagher, G.~Lombardi, P.~Gennis, Effectiveness of bystander
  cardiopulmonary resuscitation and survival following out-of-hospital cardiac
  arrest, Jama 274~(24) (1995) 1922--1925.

\bibitem{Pivac2020impact}
S.~Piva{\v{c}}, P.~Gradi{\v{s}}ek, B.~Skela-Savi{\v{c}}, The impact of
  cardiopulmonary resuscitation (cpr) training on schoolchildren and their cpr
  knowledge, attitudes toward cpr, and willingness to help others and to
  perform cpr: mixed methods research design, BMC Public Health 20 (2020)
  1--11.

\bibitem{Khanji2022Cardiopulmonary}
M.~Y. Khanji, C.~A.~A. Chahal, F.~Ricci, M.~W. Akhter, R.~S. Patel,
  Cardiopulmonary resuscitation training to improve out-of-hospital cardiac
  arrest survival: addressing potential health inequalities, European Journal
  of Preventive Cardiology 29~(17) (2022) 2275--2277.

\bibitem{Bielski2022Outcomes}
K.~Bielski, B.~W. B{\"o}ttiger, M.~Pruc, A.~Gasecka, M.~Sieminski, M.~J.
  Jaguszewski, J.~Smereka, N.~Gilis-Malinowska, F.~W. Peacock, L.~Szarpak,
  Outcomes of audio-instructed and video-instructed dispatcher-assisted
  cardiopulmonary resuscitation: a systematic review and meta-analysis, Annals
  of Medicine 54~(1) (2022) 464--471.

\bibitem{Alayrac2016Unsupervised}
J.-B. Alayrac, P.~Bojanowski, N.~Agrawal, J.~Sivic, I.~Laptev,
  S.~Lacoste-Julien, Unsupervised learning from narrated instruction videos,
  in: Proceedings of the IEEE Conference on Computer Vision and Pattern
  Recognition, 2016, pp. 4575--4583.

\bibitem{Piergiovanni2021Unsupervised}
A.~Piergiovanni, A.~Angelova, M.~S. Ryoo, I.~Essa, Unsupervised discovery of
  actions in instructional videos, arXiv preprint arXiv:2106.14733 (2021).

\bibitem{Ding2022Temporal}
G.~Ding, A.~Yao, Temporal action segmentation with high-level complex activity
  labels, IEEE Transactions on Multimedia (2022).

\bibitem{Apostolidis2021Video}
E.~Apostolidis, E.~Adamantidou, A.~I. Metsai, V.~Mezaris, I.~Patras, Video
  summarization using deep neural networks: A survey, Proceedings of the IEEE
  109~(11) (2021) 1838--1863.

\bibitem{Vishwakarma2013survey}
S.~Vishwakarma, A.~Agrawal, A survey on activity recognition and behavior
  understanding in video surveillance, The Visual Computer 29 (2013) 983--1009.

\bibitem{Jhuang2013Towards}
H.~Jhuang, J.~Gall, S.~Zuffi, C.~Schmid, M.~J. Black, Towards understanding
  action recognition, in: Proceedings of the IEEE international conference on
  computer vision, 2013, pp. 3192--3199.

\bibitem{Kong2022Human}
Y.~Kong, Y.~Fu, Human action recognition and prediction: A survey,
  International Journal of Computer Vision 130~(5) (2022) 1366--1401.

\bibitem{Liu2021Towards}
D.~Liu, Q.~Li, T.~Jiang, Y.~Wang, R.~Miao, F.~Shan, Z.~Li, Towards unified
  surgical skill assessment, in: Proceedings of the IEEE/CVF Conference on
  Computer Vision and Pattern Recognition, 2021, pp. 9522--9531.

\bibitem{Wang2020Boundary}
Z.~Wang, Z.~Gao, L.~Wang, Z.~Li, G.~Wu, Boundary-aware cascade networks for
  temporal action segmentation, in: Computer Vision--ECCV 2020: 16th European
  Conference, Glasgow, UK, August 23--28, 2020, Proceedings, Part XXV 16,
  Springer, 2020, pp. 34--51.

\bibitem{Farha2019Ms}
Y.~A. Farha, J.~Gall, Ms-tcn: Multi-stage temporal convolutional network for
  action segmentation, in: Proceedings of the IEEE/CVF conference on computer
  vision and pattern recognition, 2019, pp. 3575--3584.

\bibitem{Cheng2014Temporal}
Y.~Cheng, Q.~Fan, S.~Pankanti, A.~Choudhary, Temporal sequence modeling for
  video event detection, in: Proceedings of the IEEE conference on computer
  vision and pattern recognition, 2014, pp. 2227--2234.

\bibitem{Huang2016Connectionist}
D.-A. Huang, L.~Fei-Fei, J.~C. Niebles, Connectionist temporal modeling for
  weakly supervised action labeling, in: Computer Vision--ECCV 2016: 14th
  European Conference, Amsterdam, The Netherlands, October 11--14, 2016,
  Proceedings, Part IV 14, Springer, 2016, pp. 137--153.

\bibitem{Singh2016multi}
B.~Singh, T.~K. Marks, M.~Jones, O.~Tuzel, M.~Shao, A multi-stream
  bi-directional recurrent neural network for fine-grained action detection,
  in: Proceedings of the IEEE conference on computer vision and pattern
  recognition, 2016, pp. 1961--1970.

\bibitem{Wang2021Temporal}
D.~Wang, D.~Hu, X.~Li, D.~Dou, Temporal relational modeling with
  self-supervision for action segmentation, in: Proceedings of the AAAI
  Conference on Artificial Intelligence, Vol.~35, 2021, pp. 2729--2737.

\bibitem{Huang2020Improving}
Y.~Huang, Y.~Sugano, Y.~Sato, Improving action segmentation via graph-based
  temporal reasoning, in: Proceedings of the IEEE/CVF conference on computer
  vision and pattern recognition, 2020, pp. 14024--14034.

\bibitem{Carreira2017Quo}
J.~Carreira, A.~Zisserman, Quo vadis, action recognition? a new model and the
  kinetics dataset, in: proceedings of the IEEE Conference on Computer Vision
  and Pattern Recognition, 2017, pp. 6299--6308.

\bibitem{Vaswani2017Attention}
A.~Vaswani, N.~Shazeer, N.~Parmar, J.~Uszkoreit, L.~Jones, A.~N. Gomez,
  {\L}.~Kaiser, I.~Polosukhin, Attention is all you need, Advances in neural
  information processing systems 30 (2017).

\bibitem{Dosovitskiy2020image}
A.~Dosovitskiy, L.~Beyer, A.~Kolesnikov, D.~Weissenborn, X.~Zhai,
  T.~Unterthiner, M.~Dehghani, M.~Minderer, G.~Heigold, S.~Gelly, et~al., An
  image is worth 16x16 words: Transformers for image recognition at scale,
  arXiv preprint arXiv:2010.11929 (2020).

\bibitem{Brown2020Language}
T.~Brown, B.~Mann, N.~Ryder, M.~Subbiah, J.~D. Kaplan, P.~Dhariwal,
  A.~Neelakantan, P.~Shyam, G.~Sastry, A.~Askell, et~al., Language models are
  few-shot learners, Advances in neural information processing systems 33
  (2020) 1877--1901.

\bibitem{Radford2021Learning}
A.~Radford, J.~W. Kim, C.~Hallacy, A.~Ramesh, G.~Goh, S.~Agarwal, G.~Sastry,
  A.~Askell, P.~Mishkin, J.~Clark, et~al., Learning transferable visual models
  from natural language supervision, in: International conference on machine
  learning, PMLR, 2021, pp. 8748--8763.

\bibitem{Karaman2014Fast}
S.~Karaman, L.~Seidenari, A.~Del~Bimbo, Fast saliency based pooling of fisher
  encoded dense trajectories, in: ECCV THUMOS Workshop, Vol.~1, 2014, p.~5.

\bibitem{Kuehne2016end}
H.~Kuehne, J.~Gall, T.~Serre, An end-to-end generative framework for video
  segmentation and recognition, in: 2016 IEEE Winter Conference on Applications
  of Computer Vision (WACV), IEEE, 2016, pp. 1--8.

\bibitem{Bhattacharya2014Recognition}
S.~Bhattacharya, M.~M. Kalayeh, R.~Sukthankar, M.~Shah, Recognition of complex
  events: Exploiting temporal dynamics between underlying concepts, in:
  Proceedings of the IEEE conference on computer vision and pattern
  recognition, 2014, pp. 2235--2242.

\bibitem{Lea2017Temporal}
C.~Lea, M.~D. Flynn, R.~Vidal, A.~Reiter, G.~D. Hager, Temporal convolutional
  networks for action segmentation and detection, in: proceedings of the IEEE
  Conference on Computer Vision and Pattern Recognition, 2017, pp. 156--165.

\bibitem{zhang2018weighted}
Y.~Zhang, L.~Chen, C.~Yan, P.~Qin, X.~Ji, Q.~Dai, Weighted convolutional
  motion-compensated frame rate up-conversion using deep residual network, IEEE
  Transactions on Circuits and Systems for Video Technology 30~(1) (2018)
  11--22.

\bibitem{Chen2020Action}
M.-H. Chen, B.~Li, Y.~Bao, G.~AlRegib, Z.~Kira, Action segmentation with joint
  self-supervised temporal domain adaptation, in: Proceedings of the IEEE/CVF
  Conference on Computer Vision and Pattern Recognition, 2020, pp. 9454--9463.

\bibitem{zhang2022rcmnet}
R.~Zhang, X.~Han, Z.~Lei, C.~Jiang, I.~Gul, Q.~Hu, S.~Zhai, H.~Liu, L.~Lian,
  Y.~Liu, et~al., Rcmnet: A deep learning model assists car-t therapy for
  leukemia, Computers in Biology and Medicine 150 (2022) 106084.

\bibitem{hassan2022neuro}
M.~Hassan, H.~Guan, A.~Melliou, Y.~Wang, Q.~Sun, S.~Zeng, W.~Liang, Y.~Zhang,
  Z.~Zhang, Q.~Hu, et~al., Neuro-symbolic learning: Principles and applications
  in ophthalmology, arXiv preprint arXiv:2208.00374 (2022).

\bibitem{hassan2023retinal}
M.~Hassan, H.~Zhang, A.~A. Fateh, S.~Ma, W.~Liang, D.~Shang, J.~Deng, Z.~Zhang,
  T.~K. Lam, M.~Xu, et~al., Retinal disease projection conditioning by
  biological traits, Complex \& Intelligent Systems (2023) 1--15.

\bibitem{bhardwaj2022machine}
V.~Bhardwaj, A.~Sharma, S.~V. Parambath, I.~Gul, X.~Zhang, P.~E. Lobie, P.~Qin,
  V.~Pandey, Machine learning for endometrial cancer prediction and
  prognostication, Frontiers in Oncology 12 (2022) 852746.

\bibitem{guan2023prevalence}
J.~Guan, Y.~Zhu, Q.~Hu, S.~Ma, J.~Mu, Z.~Li, D.~Fang, X.~Zhuo, H.~Guan, Q.~Sun,
  et~al., Prevalence patterns and onset prediction of high myopia for children
  and adolescents in southern china via real-world screening data:
  Retrospective school-based study, Journal of Medical Internet Research 25
  (2023) e39507.

\bibitem{zhang2022ai}
L.~Zhang, Z.~Lei, C.~Xiao, Z.~Du, C.~Jiang, X.~Yuan, Q.~Hu, S.~Zhai, L.~Xu,
  C.~Liu, et~al., Ai-boosted crispr-cas13a and total internal reflection
  fluorescence microscopy system for sars-cov-2 detection, Frontiers in Sensors
  3 (2022) 1015223.

\bibitem{liu2022mixed}
Y.~Liu, L.~Lian, E.~Zhang, L.~Xu, C.~Xiao, X.~Zhong, F.~Li, B.~Jiang, Y.~Dong,
  L.~Ma, et~al., Mixed-unet: Refined class activation mapping for
  weakly-supervised semantic segmentation with multi-scale inference, Frontiers
  in Computer Science 4 (2022) 1036934.

\bibitem{chen2021accelerated}
X.~Chen, B.~Li, S.~Jiang, T.~Zhang, X.~Zhang, P.~Qin, X.~Yuan, Y.~Zhang,
  G.~Zheng, X.~Ji, Accelerated phase shifting for structured illumination
  microscopy based on deep learning, IEEE Transactions on Computational Imaging
  7 (2021) 700--712.

\bibitem{chen2023crispr}
Q.~Chen, I.~Gul, C.~Liu, Z.~Lei, X.~Li, M.~A. Raheem, Q.~He, Z.~Haihui,
  E.~Leeansyah, C.~Y. Zhang, et~al., Crispr--cas12-based field-deployable
  system for rapid detection of synthetic dna sequence of the monkeypox virus
  genome, Journal of Medical Virology 95~(1) (2023) e28385.

\bibitem{Yi2021Asformer}
F.~Yi, H.~Wen, T.~Jiang, Asformer: Transformer for action segmentation, arXiv
  preprint arXiv:2110.08568 (2021).

\bibitem{Du2022Do}
D.~Du, B.~Su, Y.~Li, Z.~Qi, L.~Si, Y.~Shan, Do we really need temporal
  convolutions in action segmentation, arXiv preprint arXiv:2205.13425 2 (2022)
  13.

\bibitem{Wang2022Cross}
J.~Wang, Z.~Wang, S.~Zhuang, H.~Wang, Cross-enhancement transformer for action
  segmentation, arXiv preprint arXiv:2205.09445 (2022).

\bibitem{Ishikawa2021Alleviating}
Y.~Ishikawa, S.~Kasai, Y.~Aoki, H.~Kataoka, Alleviating over-segmentation
  errors by detecting action boundaries, in: Proceedings of the IEEE/CVF winter
  conference on applications of computer vision, 2021, pp. 2322--2331.

\bibitem{Jia2021Scaling}
C.~Jia, Y.~Yang, Y.~Xia, Y.-T. Chen, Z.~Parekh, H.~Pham, Q.~Le, Y.-H. Sung,
  Z.~Li, T.~Duerig, Scaling up visual and vision-language representation
  learning with noisy text supervision, in: International Conference on Machine
  Learning, PMLR, 2021, pp. 4904--4916.

\bibitem{Wang2021Actionclip}
M.~Wang, J.~Xing, Y.~Liu, Actionclip: A new paradigm for video action
  recognition, arXiv preprint arXiv:2109.08472 (2021).

\bibitem{Li2022Bridge}
M.~Li, L.~Chen, Y.~Duan, Z.~Hu, J.~Feng, J.~Zhou, J.~Lu, Bridge-prompt: Towards
  ordinal action understanding in instructional videos, in: Proceedings of the
  IEEE/CVF Conference on Computer Vision and Pattern Recognition, 2022, pp.
  19880--19889.

\bibitem{stein2013combining}
S.~Stein, S.~J. McKenna, Combining embedded accelerometers with computer vision
  for recognizing food preparation activities, in: Proceedings of the 2013 ACM
  international joint conference on Pervasive and ubiquitous computing, 2013,
  pp. 729--738.

\bibitem{kuehne2014language}
H.~Kuehne, A.~Arslan, T.~Serre, The language of actions: Recovering the syntax
  and semantics of goal-directed human activities, in: Proceedings of the IEEE
  conference on computer vision and pattern recognition, 2014, pp. 780--787.

\bibitem{fathi2011learning}
A.~Fathi, X.~Ren, J.~M. Rehg, Learning to recognize objects in egocentric
  activities, in: CVPR 2011, IEEE, 2011, pp. 3281--3288.

\bibitem{Lea2016Segmental}
C.~Lea, A.~Reiter, R.~Vidal, G.~D. Hager, Segmental spatiotemporal cnns for
  fine-grained action segmentation, in: Computer Vision--ECCV 2016: 14th
  European Conference, Amsterdam, The Netherlands, October 11-14, 2016,
  Proceedings, Part III 14, Springer, 2016, pp. 36--52.

\bibitem{Li2020MS}
S.-J. Li, Y.~AbuFarha, Y.~Liu, M.-M. Cheng, J.~Gall, Ms-tcn++: Multi-stage
  temporal convolutional network for action segmentation, IEEE Transactions on
  Pattern Analysis and Machine Intelligence (2020).

\bibitem{Lea2016Learning}
C.~Lea, R.~Vidal, G.~D. Hager, Learning convolutional action primitives for
  fine-grained action recognition, in: 2016 IEEE international conference on
  robotics and automation (ICRA), IEEE, 2016, pp. 1642--1649.

\end{thebibliography}

\end{document}